\documentclass[11pt]{article}
\usepackage[utf8]{inputenc}
\usepackage[T1]{fontenc}
\usepackage{graphicx}
\usepackage{longtable}
\usepackage{wrapfig}
\usepackage{rotating}
\usepackage[normalem]{ulem}
\usepackage{amsmath}
\usepackage{amssymb}
\usepackage{capt-of}
\usepackage{hyperref}
\usepackage{coling}
\usepackage{subcaption}
\usepackage{times}
\usepackage{booktabs} % For better table formatting
\usepackage{microtype}
\usepackage{inconsolata}
\usepackage{soul}
\usepackage{todonotes}

\usepackage{enumitem}

\author{
    Nouran Khallaf, Carlo Eugeni, Serge Sharoff \\[2ex]  % Author names
    \small University of Leeds, UK\\
    \small \texttt{N.Khallaf, C.Eugeni, S.Sharoff @leeds.ac.uk} % Emails
}

\date{For Writing Aid \url{https://sites.google.com/view/wraicogs1/home/background-and-topics}}
\title{Reading Between the Lines: \\A dataset and a study on why some texts are tougher than others}
\hypersetup{
 pdfauthor={Nouran Khallaf, Carlo Eugeni, Serge Sharoff},
 pdftitle={Reading Between the Lines: A dataset and a study on why some texts are tougher than others},
 pdfkeywords={},
 pdfsubject={},
 pdfcreator={Emacs 29.3 (Org mode 9.6.15)}, 
 pdflang={English}}
\begin{document}

\maketitle
\setlist{itemsep=5pt,parsep=0pt}

\begin{abstract}
Our research aims at better understanding what makes a text difficult to read for specific audiences with intellectual disabilities, more specifically, people who have limitations in cognitive functioning,
such as reading and understanding skills, an IQ below 70, and challenges in conceptual domains.  We introduce a scheme for the annotation of difficulties which is based on empirical research in psychology as well as on research in translation studies. 
The paper describes the annotated dataset, primarily derived from the parallel texts (standard English and Easy to Read English translations) made available online. we fine-tuned four different pre-trained transformer models to perform the task of multiclass classification to predict the strategies required for simplification. We also investigate the possibility to interpret the decisions of this language model when it is aimed at predicting the difficulty of sentences. The resources are available from https://github.com/Nouran-Khallaf/why-tough
\end{abstract}

\section{Introduction}
\label{sec:org777e9e7}
The Universal Declaration of Human Rights, in its Article 19,
affirms everyone's right to seek and receive information. Similarly, Article 21 of the UN Convention on the Rights of Persons with Disabilities underscores the need for accessible formats, ensuring that individuals with disabilities can access public information without additional cost. For people with intellectual
disabilities---those with limitations in cognitive functioning,
including difficulties in reading and understanding, an IQ below 70, and
challenges in conceptual domains \cite{aaidd_id_faqs}---language simplification
is crucial for ensuring accessibility and equality, making it essential
for them to fully enjoy their human rights.

Text Simplification (TS) research aims to make text easier to read while preserving its meaning and key information \cite{saggion2017simplification}. Earlier studies involved lexical, syntactic and semantic modifications, while modern research benefits from the use of Large Language Models (LLMs), with still unclear cost-to-performance benefits, as they do not outperform smaller Pre-trained Language Models (PLMs), such as BERT, on text classification tasks \cite{edwards24incontext}.

Computational studies often overlook insights from translation studies, particularly the various strategies proposed \citep{vinay71, newmark1988textbook, chesterman97, zabalbeascoa2000solutions, molina2002techniques, gambier2006audiovisual}, focusing on the systematic processes involved in translating a source text into a target text across languages. Translation studies provide a complementary lens by examining strategies used in intralingual translation, where a source text is converted into a target text in the same language. \citet[82]{eugeni23} argue that such shifts often achieve full correspondence between source and target texts. Of particular relevance are two types of intralingual translation. Diamesic Translation involves shifting communication modes (e.g., spoken to written) while retaining the same language \cite{eugeni2020}.

\textbf{Diastratic Translation}, on the other hand, involves register shifts within the same language, such as from Standard English (SE) to Easy to Read (E2R) English, i.e.~the variation of language that is easy to read and understand for people with reading difficulties, including people with intellectual disabilities, people with little command of the language, people with poor literacy and so forth \cite{europe09,bernabe2017}. 
Compared to standard language E2R language is a simplified version for the sake of readability for specific audiences  \cite{bernabe2017}. As a result, it forms the foundation of diverse and adaptable translation strategies designed to make information accessible to people with intellectual disabilities.

Previous studies in text simplification have primarily focused on lexical simplification, where individual words or phrases are simplified without considering the broader sentence structure or context. For instance, \citet{saggion2015lexical} developed datasets and tools specifically tailored for lexical simplification tasks, emphasising word-level transformations. While this approach has proven effective for specific applications, it often overlooks the interplay between lexical and syntactic features within a sentence.

Other notable resources, such as the ASSET corpus \cite{alva2020asset}, have focused on sentence simplification but rely on predefined, fine-grained operations at the word or phrase level. Similarly, corpora like WikiLarge \cite{zhang2017wikilarge} offer paired datasets for simplification but lack explicit annotations for the strategies applied during simplification. These resources are invaluable for training machine learning models but are limited in their ability to capture a comprehensive view of the simplification process.

In contrast to the resources mentioned above, our dataset adopts a holistic approach to sentence simplification, focusing on sentence-level transformations that encompass lexical, syntactic, and semantic changes, while focusing on the reason to make these changes. Unlike lexical simplification datasets, which isolate individual words or phrases, our dataset explicitly annotates entire sentences with six predefined categories representing diverse simplification strategies. This allows for better understanding of the simplification process, capturing how different strategies interact within a sentence to enhance its readability and accessibility.

Furthermore, by annotating SE and E2R sentence pairs, our dataset provides a unique resource for exploring context-sensitive simplification strategies. This makes it particularly valuable for tasks that require an integrated understanding of sentence-level transformations.

This study explores strategies to make information more accessible through text simplification. Our contributions concern:
(1) the development of an extended taxonomy of translation strategies that integrates insights from Text Simplification research, (2) the annotation of a parallel corpus of complex and simplified texts sourced from diverse public services in Scotland (see Section \ref{secDataset}), (3) the investigation of setting to train transformer-based models to predict the application of specific simplification strategies, and (4) an investigation into interpretability of their predictions using Explainable AI (XAI) techniques to explain the model’s decision-making process. While Large Language Models (LLMs) demonstrate impressive performance, their \textit{``black-box''} nature often makes it challenging to understand their predictions. To address this, we employ Integrated Gradients \cite{sundararajan17ig}, an XAI method grounded in axiomatic attribution principles.
IG identifies the most influential words in the input by analysing gradient variation. By aligning these attributions with human judgments, we enhance the interpretability of the model and build trust in its application.

\begin{table*}[!t]
\caption{Snapshot of Scottish Government Dataset Statistics \label{tabCorpus}}
\centering
\small
\begin{tabular}{l|r|rrr|rrr}
\textbf{Source} & \textbf{\#Texts} & \textbf{Complex} &  &  & \textbf{Simple} &  & \\[0pt]
 &  & \#Words & \#Sentences & IQR & \#Words & \#Sentences & IQR \\[0pt]

Health & 21 & 183677 & 7258 & (15.0-31.0) & 30253 & 1519 & (10.0-21.0) \\[0pt]
Public info & 4 & 12217 & 527 & (12.0-30.5) & 3378 & 217 & (9.0-18.0) \\[0pt]
Politics & 9 & 113412 & 4824 & (15.0-29.0) & 12474 & 832 & (9.0-17.0) \\[0pt]
\hline
Data selection & -- & 4166 & 155 & (12-27) & 3259 & 161 & (9-20) \\[0pt]
\hline

\end{tabular}
\end{table*}

\section{Dataset}
\label{secDataset}

The original corpus consists of over 76 parallel texts, primarily sourced from the Scottish care service, political manifestos for the 2024 UK general election, and newsletters from the national charity Disability Equality Scotland. These texts span a diverse range of topics, including health care services, environmental policies, the legal system, waste management, disability advocacy, and linguistic accessibility.

%\subsection{Dataset Preparation and Preprocessing}
%%As shown in Table \ref{tabCorpus}, the complex version contains a total of 4,166 words across 206 sentences, with an interquartile range (IQR) of 12–27 words per sentence. In contrast, while the simplified version has a reduced total word count of 3,259 words, it includes a slightly higher number of sentences at 210, with a narrower IQR of 9–20 words per sentence. 

Table \ref{tabCorpus} compares information about the original documents ("complex") with their simplified versions in terms of the number of words and sentences in each corpus part as well as the Inter-Quartile Range of the sentence lengths measured in words.  The overall word count and average sentence length have significantly decreased for the simplified version compared to the complex texts, in spite of some of the strategies aimed at explanation and sentence splitting. This increase in the number of sentences, coupled with the reduction in word count, reflects a structural adjustment typical of simplification strategies, which often involves breaking down longer sentences into shorter, more accessible ones to enhance readability.

\autoref{tabStrategies} lists the general strategies for simplification, while \autoref{tabStrategiesSimple} lists the fine-grained annotation categories used for annotation. A detailed breakdown of macro typology frequencies within their corresponding main strategies showcases the distribution of techniques and methods employed to simplify texts. The prominence of semantic and explanation categories reflects a strong emphasis on clarity and enhancing reader accessibility.

\begin{table*}[htbp]
\caption{Macro-Strategies and Corresponding Strategies for Simplification \label{tabStrategies}}
\centering
\footnotesize % Reduce font size
\begin{tabular}{|l|p{11cm}|}
\hline
\textbf{Macro-Strategy} & \textbf{Strategies} \\
\hline
\textbf{Transcription} & No simplification needed. \\
\hline
\textbf{Synonymy} &
\textbf{Pragmatic:} Acronyms spelled out; Proper names to common names; Contextual synonyms made explicit. \newline
\textbf{Semantic:} Hyperyms; Hyponyms; Stereotypes. \newline
\textbf{Grammatical:} Negative to positive sentences; Passive to active sentences; Pronouns to referents; Tenses simplified. \\
\hline
\textbf{Explanation} &
Words given for known; Expressions given for known; Tropes explained; Schemes explained; Deixis clarified; Hidden grammar made explicit; Hidden concepts made explicit. \\
\hline
\textbf{Syntactic Changes} &
Word $\rightarrow$ Group; Word $\rightarrow$ Clause; Word $\rightarrow$ Sentence; Group $\rightarrow$ Word; Group $\rightarrow$ Clause; Group $\rightarrow$ Sentence; Clause $\rightarrow$ Word; Clause $\rightarrow$ Group; Clause $\rightarrow$ Sentence; Sentence $\rightarrow$ Word; Sentence $\rightarrow$ Group; Sentence $\rightarrow$ Clause. \\
\hline
\textbf{Transposition} &
Nouns for things, animals, or people; Verbs for actions; Adjectives for nouns; Adverbs for verbs. \\
\hline
\textbf{Modulation} &
Text-level linearity; Sentence-level linearity: Chronological order of clauses; Logical order of complements. \\
\hline
\textbf{Anaphora} & Repetition replaces synonyms. \\
\hline
\textbf{Omission} &
Useless elements: Nouns; Verbs; Complements; Sentences. \newline
Rhetorical constructs; Diamesic elements. \\
\hline
\textbf{Illocutionary Change} & Implicit meaning made explicit. \\
\hline
\textbf{Compression} &
Grammatical constructs simplified; Rhetorical constructs simplified. \\
\hline
\end{tabular}
\end{table*}

\begin{table*}[t]
\caption{A subset of strategies in dataset annotations and their annotation labels \label{tabStrategiesSimple}}
\centering
\small
\begin{tabular}{|p{3cm}|p{11cm}|}
\hline
\textbf{Macro-Strategy} & \textbf{Strategies} \\
\hline
\textbf{Omission} & OmiSen, OmiWor, OmiClau, OmiRhet (on the level of sentences, words, clauses or rhetorical structures)\\
\hline
\textbf{Compression} & SinGram, SimGram, SinSem, SinPrag \\
\hline
\textbf{Explanation} & ExplWor, ExplCont, ExplExpr, HidCont, HidGram, WordExpl \\
\hline
\textbf{Syntactic Changes} & SynChange, Clause2Word, WordsOrder, GroupOrder, LinearOrderSen, LinearOrderCla \\
\hline
\textbf{Substitution} & Anaph, SynSem, SemStereo \\
\hline
\textbf{Transposition} & TranspNoun \\
\hline
\textbf{Modulation} & ModInfo \\
\hline
\end{tabular}
\end{table*}

In the field of Translation Studies, many taxonomies have been developed to identify the strategies professional translators apply when producing a target text. Most of these strategies have been developed in the field of interlingual translation, first from a written text into another written text \cite{nida64,vinay71,chesterman97,molina02}, and then from a spoken text into a written text \cite{gottlieb92,lambert96,ivarsson98,lomheim95,kovacic00}. The study of intralingual translation strategies is relatively more recent and mainly focuses on \textbf{Diamesic Translation} \cite{neves05,eugeni07,brumme08,gambier10,eugeni23}. Rarer is the number of authors who have tried to define strategies for the translation of written texts within the same language \cite{zethsen09,ersland14}. To our knowledge, only \citet{hansen-schirra20} and \citet{maass20} have addressed intralingual translation practices into E2R.

However, none of these taxonomies completely satisfy our need to account for all the simplification strategies we identified in our corpus, as too little detail was provided. The opposite happens in the completely different field of Automatic Text Simplification (ATS), where details are, instead, provided. Here, the focus of typologies is on linguistic descriptions and string edits. A significant contribution in ATS has been provided by \citet{cardon2022}, whose typology essentially focuses on operations that mainly deal with adding, deleting, replacing, and moving words. However, texts translated in E2R language clearly show that professionals in the field apply many more operations that pertain to the field of pragmatics and semiotics, focused on how concepts are distributed and or explained to help the user understand them. It is in this context that this section will try to illustrate the annotation framework that we have developed and used in this study. Because the form of translation we are focussing on in this paper is diastratic (from SE to E2R), we used Inclusion Europe’s pioneering guidelines \citet{europe09} as a basis for our annotation framework, which was then used to identify the strategies used in our corpus.

The principle of Inclusion Europe's guidelines is language
simplification, further subdivided into three levels: lexical,
syntactical, and semantic. The lexical level mainly focuses on the use of
nouns, verb tenses, adjectives, and adverbs. In particular, the guidelines require to only use basic vocabulary words. For the English language, the Basic Vocabulary \cite{ogden1932basic} – that has evolved into projects like Voice of America's Word Book of around 1500 words – contains 850 commonly used word roots, like thing, do, good, or very. The syntactical level mainly focuses on the use of the
order of words and clauses in a sentence, and that of sentences in the
text. In particular, the guidelines require to only use a
(chrono-)logically linear word, clause, and sentence order. The semantic
level mainly focuses on the distribution of concepts in the text. In
particular, the guidelines require one concept per sentence.
\emph{Information for all} also add other pieces of information, like
the use of pictograms to reinforce the information provided in the text.
% , a validation team composed of people with various types of reading
% difficulties that can approve the readability of the text, and the use
% of the E2R logo. 
However, these will not be considered in the present study.

\begin{figure}
    \centering
    \includegraphics[width=0.7\linewidth]{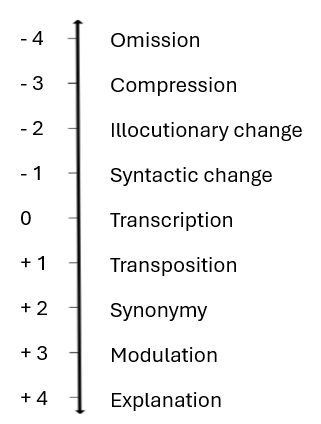}
    \caption{Diastratic Translation Strategies distributed along a continuum, from most deduction of text (-4) to most addition of text (+4)}
    \label{figDiastratic}
\end{figure}

Based on these principles, and a qualitative analysis of the illustrated
corpus, we came up with the following nine macro-strategies, that easily
adapt to our heterogenous corpus. Macro-strategies are further
subdivided into strategies and micro-strategies. The macro-strategies
have been thought as points in a continuum between two poles: those
resulting in most addition of text (explanation) to those resulting in
the most deduction of text (omission), the middle being constituted by
transcription, with no addition or deduction of text (\autoref{figDiastratic}).
Examples are taken from our corpus.

1. \textit{Explanation}, which includes the explicitation of hidden grammar or
content (e.g. ``wherever they live'' $\to$ ``wherever they live in
Scotland''), or the explanation of a word or expression that is given
for known (e.g. ``\textbf{co-design} services with people with
experience of accessing and delivering them'' $\to$
``\textbf{co-design} services with people who use or work in them and
their carers. \textbf{Co-design means} you can share your ideas and
experiences with us.'').

2. \textit{Modulation} is the distribution of information in a linear order in
the text and in a sentence, according to the principle that one sentence
should contain one piece of information only.
This means that one sentence is turned into more sentences (e.g. ``He
joins in community activities as much as possible, supported by his
assistants and his family.'' $\to$ ``He likes to take part in
activities where he can meet people. He gets support from his assistants
and his family.'') or words are redistributed within the sentence (e.g.
``The NCS will make collaboration and \textbf{information sharing}
between these services easier'' $\to$ ``The NCS will make
working together and \textbf{sharing information} easier for
services.'').

3. \textit{Synonymy}, whereby a complex, technical, or abstract word is replaced
by a more common and concrete one. Synonymy includes pragmatic synonyms
that depend on the context (e.g. ``sir Keir Starmer'' $\to$
``the new Prime Minister''), as well as semantic synonyms (e.g.
``conversation'' $\to$ ``talk''), and grammatical synonyms
(e.g. ``The money does not have to be paid back'' $\to$ ``You
do not have to pay the money back'') that depend on grammar.

4. \textit{Transposition}, or word class change, whereby the class of a word is
changed depending on the principle that nouns should ideally stand for
things, animals, or people, and verbs stand for actions (e.g. ``our aim
is'' $\to$ the Scottish Government wants'').

5. \textit{Transcript}, by which the words of the source text are left unchanged
because no simplification is needed (e.g. ``I love music'').

6. \textit{Syntactic change}, whereby a word, group, clause, or sentence is
turned into one of the other three syntactic levels (e.g. citizens →
people living in Scotland).

7. \textit{Illocutionary change}, by which what is implied is said (e.g. ``I like
to say that we, the dancers, must gather information about our body's
library $\to$ ``The dancers must know their own body.'').

8. \textit{Compression} of grammatical or semantic constructs (e.g. ``The
moderator asks questions and shows slides, pictures or videos \textbf{to
guide the group}'' $\to$ ``The moderator asks questions and
shows slides, pictures, or videos \textbf{to the group}'').

%% \tdh{Carlo}{is this from our corpus?}
9. \textit{Omission} of rhetorical or diamesic constructs (e.g. ``I was nervous,
\textbf{of course}, but it was interesting and fun!'' $\to$ ``I
was worried, but it was interesting and fun!''), or of what is
considered useless for understanding an idea at the noun, verb,
complement or sentence level (e.g. ``\textbf{Sir Keir Rodney Starmer KCB
KC} is a British politician'' $\to$ \textbf{Starmer} is a
British politician'').

\section{Classification Model: Multiclass Text Classification with Transformers}

This experiment investigates the application of pre-trained transformer-based models for multiclass text classification, focusing on the prediction of simplification strategies need to simplify the respective SE sentences. 
% By addressing challenges such as class imbalance, the experiment aims to provide a robust framework for typology prediction using seven predefined categories.

For this experiment, seven categories were manually annotated for a selection of 155 complex sentences and their 161 corresponding simplified sentences, randomly selected from various texts see Table \ref{tabCorpus}. The seven categories—\textit{Explanation}, \textit{Grammatical Adjustments}, \textit{Modulation}, \textit{Omission}, \textit{Substitution}, \textit{Transposition}, and \textit{Syntactic Changes}—were applied to ensure coverage of multiple topics and simplification strategies. This selection was designed to create a balanced dataset that represents diverse contexts and simplification strategies. These labels are not hierarchical but independent categories reflecting distinct simplification strategies.

The annotation process consisted of a first analysis of the parallel texts, and a review of the existing typologies used to illustrate translation operations, both in the field of computational linguistics and translation studies. Thanks to these contributions, we came to the definition of the typology provided in Table \ref{tabCorpus}. 

%% \todo{carlo in his words he said that it was done by two?? but we can not have this }Due to resource constraints, the annotations were performed by a single annotator with expertise in text simplification and translation studies. As such, inter-annotator agreement metrics, which are standard for measuring annotation reliability, were not applicable in this study. While this ensured consistency in the application of categories, the absence of multiple annotators limits the ability to evaluate annotation reliability objectively.

%The training dataset consists of Standard English sentences paired with their E2R counterparts, annotated with fine-grained simplification strategies. To streamline classification, these fine-grained strategies were mapped to broader macro-categories based on a predefined hierarchical structure, simplifying the labels while preserving their semantic distinctions. Rows containing missing values in key columns, such as Standard English, E2R version, or Typology, were removed to ensure data quality. The paired texts were then combined into single input sequences using a separator token (\texttt{[SEP]}), allowing the model to capture relationships between the two inputs.  Finally, the dataset was split into training and validation subsets, allocating 161 samples for training and 41 samples for validation. This stratified split ensured that the distribution of categories remained consistent across both subsets.

The training dataset consists of Standard English sentences paired with their simplified counterparts. Each simplified counterpart was designed to include precisely one simplification strategy, where a single complexity was restored to its original form. This design ensures that the relationship between a sentence and its simplified version highlights specific simplification strategies, allowing the model to associate each sentence with different parts of the complexity being resolved. To streamline classification, these fine-grained simplification strategies were mapped to broader macro-categories based on a predefined hierarchical structure, simplifying the labels while preserving their semantic distinctions.

%% Finally, the dataset was split into training and validation subsets, allocating 121 samples for training and 34 samples for validation. 

\subsection{Model and Training Procedure}

We fine-tuned four different pre-trained transformer models to perform the task of multiclass classification, predicting the most likely simplification typology for each Standard English sentence.

\paragraph{Cross-Validation and Early Stopping}
We employed \textit{Stratified 5-Fold Cross-Validation} to ensure robust evaluation and generalizability. The dataset was split into four folds, maintaining the proportional distribution of typologies across training and validation sets. For each fold, the model was trained on four folds and validated on the remaining fold, and this process was repeated for all five folds. The validation results were averaged across all folds to compute the final scores. 

We used early stopping, where training was terminated if the validation loss did not improve for the patience period. This ensured efficient use of resources while retaining the best model.

\paragraph{Class Imbalance and Weighted Loss Function}
Class imbalance in the dataset, where certain typologies were underrepresented, posed a challenge during training. To address this, we utilised a \textit{weighted cross-entropy loss function}. Class weights were calculated based on the inverse frequency of each category:

\begin{equation}
w_c = \frac{1}{\text{freq}_c} \cdot \frac{N}{2},
\end{equation}

where \(w_c\) is the weight assigned to class \(c\), \(\text{freq}_c\) is the frequency of class \(c\), and \(N\) is the total number of samples. This approach ensured that underrepresented classes contributed more significantly to the overall loss, improving the model's ability to predict these minority classes.

\paragraph{Gradient Clipping}
Additionally, gradient clipping was applied during training to stabilise the optimisation process. Gradient clipping limits the maximum value of gradients during backpropagation, preventing excessively large updates to model parameters that could destabilise training or lead to divergence. Following best practices in training transformer-based models \cite{devlin2019bert}, we used a clipping threshold of \(1.0\). This ensures that gradients exceeding the threshold are scaled proportionally while gradients below the threshold remain unchanged. Mathematically, gradient clipping can be expressed as:

\begin{equation}
g_{\text{clipped}} = \min\left(g, \frac{g_{\text{threshold}}}{\|g\|}\right),
\end{equation}

\noindent
where \(g\) represents the original gradient vector, \(g_{\text{threshold}}\) is the clipping threshold (in this case, \(1.0\)), and \(\|g\|\) is the norm of the gradient vector. Gradient clipping ensures consistent updates to model parameters, improving training stability.

\paragraph{Transformer Models and Training Configuration}
Each of the four transformer models was fine-tuned for the task, using the same training configuration. 
The hyperparameters and training configuration are summarised in Table~\ref{tab:hyperparameters}.

%% Tokenisation was performed with each model’s pre-trained tokenizer, which truncated and padded input sequences to a maximum length of 512 tokens. Training was conducted for a maximum of 20 epochs, with early stopping employed to prevent overfitting.

\begin{table}[h!]
\caption{Hyperparameters and Training Configuration}
\label{tab:hyperparameters}
\centering
\small
\begin{tabular}{p{0.45\linewidth}|p{0.45\linewidth}}
\hline
\textbf{Parameter}            & \textbf{Value}                \\
\hline
Pre-trained Models            & \texttt{bert-large-cased}, \\
                              & \texttt{bert-base-multilingual} \\
                              & \hspace{5em}\texttt{cased}, \\
                              & \texttt{roberta-base}, \\
                              & \texttt{roberta-large}        \\
Max\_Sequence\_Length         & 512 tokens                   \\
Tokenisation                  & Pre-trained tokenizer        \\
Loss Function                 & Weighted Cross-Entropy Loss  \\
Class Weights                 & Inverse frequency of categories \\
Gradient\_Clipping Threshold  & 1.0                          \\
Learning Rate                 & \(5 \times 10^{-6}\)         \\
Batch Size                    & 8                            \\
Weight Decay                  & 0.01                         \\
Number of Epochs              & Up to 20 (early stopping)    \\
Cross-Validation              & Stratified 5-Fold            \\
Early Stopping Patience       & 3 epochs                     \\
GPU  & NVIDIA Tesla T4 ((15 GB memory)), \& Occasionally P100/V100 \\

\hline
\end{tabular}
\end{table}

\subsection{Evaluation Metrics and Results}

To evaluate the performance of our models, we first established a baseline using a majority-class prediction approach. This naive model assigns the most frequent class, "Explanation," to all samples. The baseline achieved an accuracy of 24.5\% and a weighted F1-score of 9.6\%. Its macro F1-score, reflecting performance across all classes equally, was only 5.6\%, highlighting its inability to handle class imbalance effectively. These results demonstrate the need for a robust machine learning model to capture the nuances of the dataset.

In contrast, our fine-tuned model (mBERT) significantly outperformed the baseline. It achieved an accuracy of 70\% and a weighted F1-score of 72\%. The macro F1-score of the multilingual model reached 65\%, reflecting its ability to generalise across minority classes.

\begin{figure}[h!]
    \centering
    \includegraphics[width=0.5\textwidth]{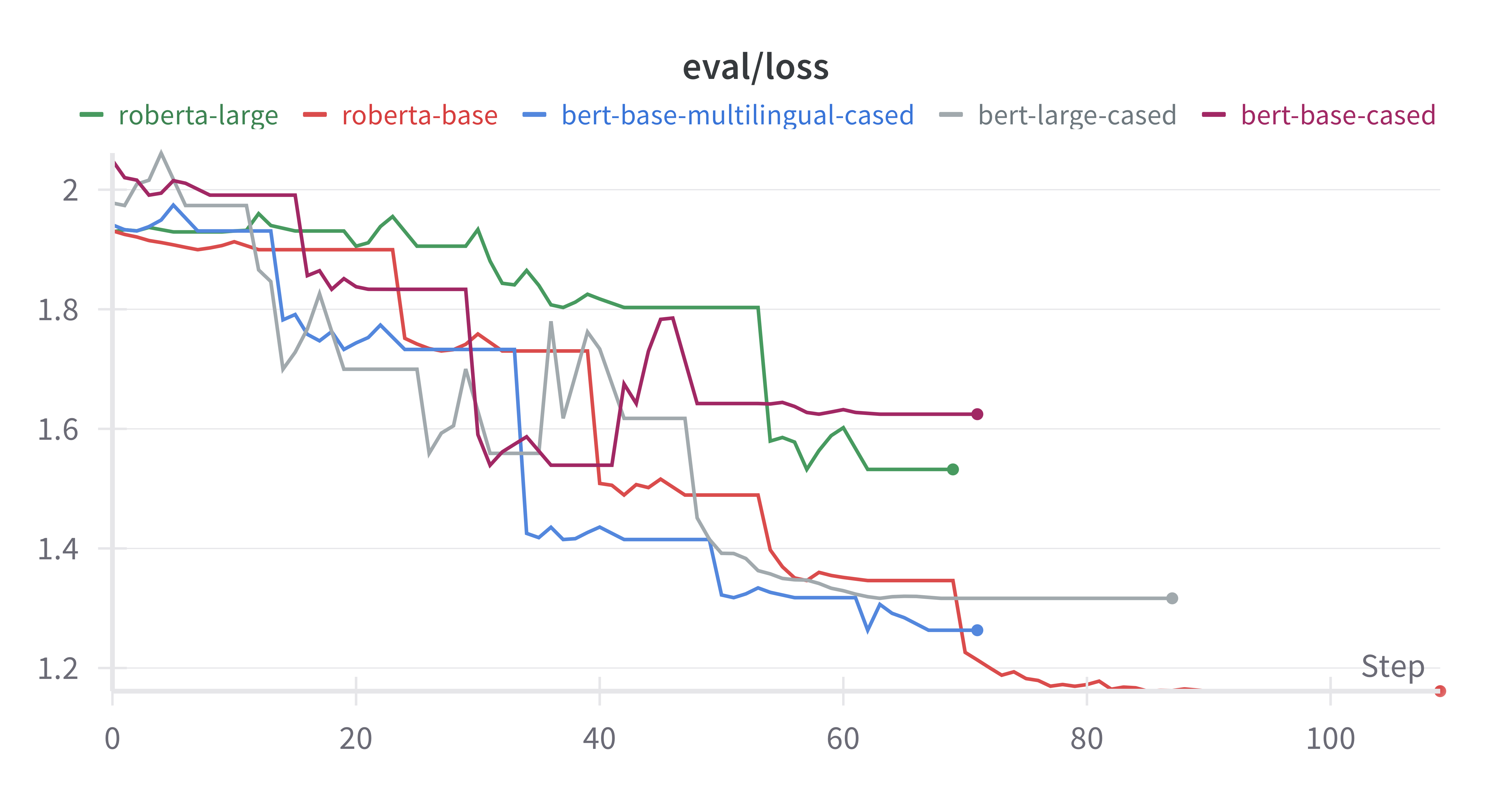}
    \caption{Evaluation Loss Progression During Training}
    \label{fig:eval_loss}
\end{figure}

In contrast, the other models demonstrated varying degrees of performance. While \texttt{roberta-base} and \texttt{roberta-large} produced reasonable results for specific classes, their overall weighted F1-scores lagged behind at 0.52 and 0.50, respectively. Similarly, \texttt{bert-large-cased} delivered moderate results with a weighted F1-score of 0.50 and accuracy of 0.53. The instability observed in the training of \texttt{roberta-base} and \texttt{roberta-large}, as evident from Figure~\ref{fig:eval_loss}, likely contributed to their lower overall scores.

The \texttt{mBERT} model excelled in identifying simplification strategies for the \textit{Explanation} (F1-score: 0.93), \textit{Substitution} (F1-score: 0.67), and \textit{Syntactic Changes} (F1-score: 0.80) categories. These results highlight its ability to capture the relationships inherent in these categories. However, underrepresented classes like \textit{Grammatical Adjustments} and \textit{Transposition} remained challenging for all models, with low F1-scores across the board. This indicates the need for a more balanced dataset.

%% The \texttt{mBERT} model not only demonstrated superior performance but also showed exceptional computational efficiency. It completed training in just 300.55 seconds, making it the second fastest model after \texttt{roberta-base}, which trained in 219.30 seconds. By comparison, \texttt{roberta-large} required significantly more time at 664.41 seconds, highlighting the computational cost of larger models.

Figure~\ref{fig:eval_loss} illustrates the evaluation loss progression during training, where the \texttt{mBERT} model exhibited a smooth and consistent reduction in loss, indicating stable convergence. In contrast, \texttt{roberta-base} and \texttt{roberta-large} displayed oscillatory behavior, suggesting instability in their training dynamics.

\begin{figure}[h!]
    \centering
    \includegraphics[width=0.5\textwidth]{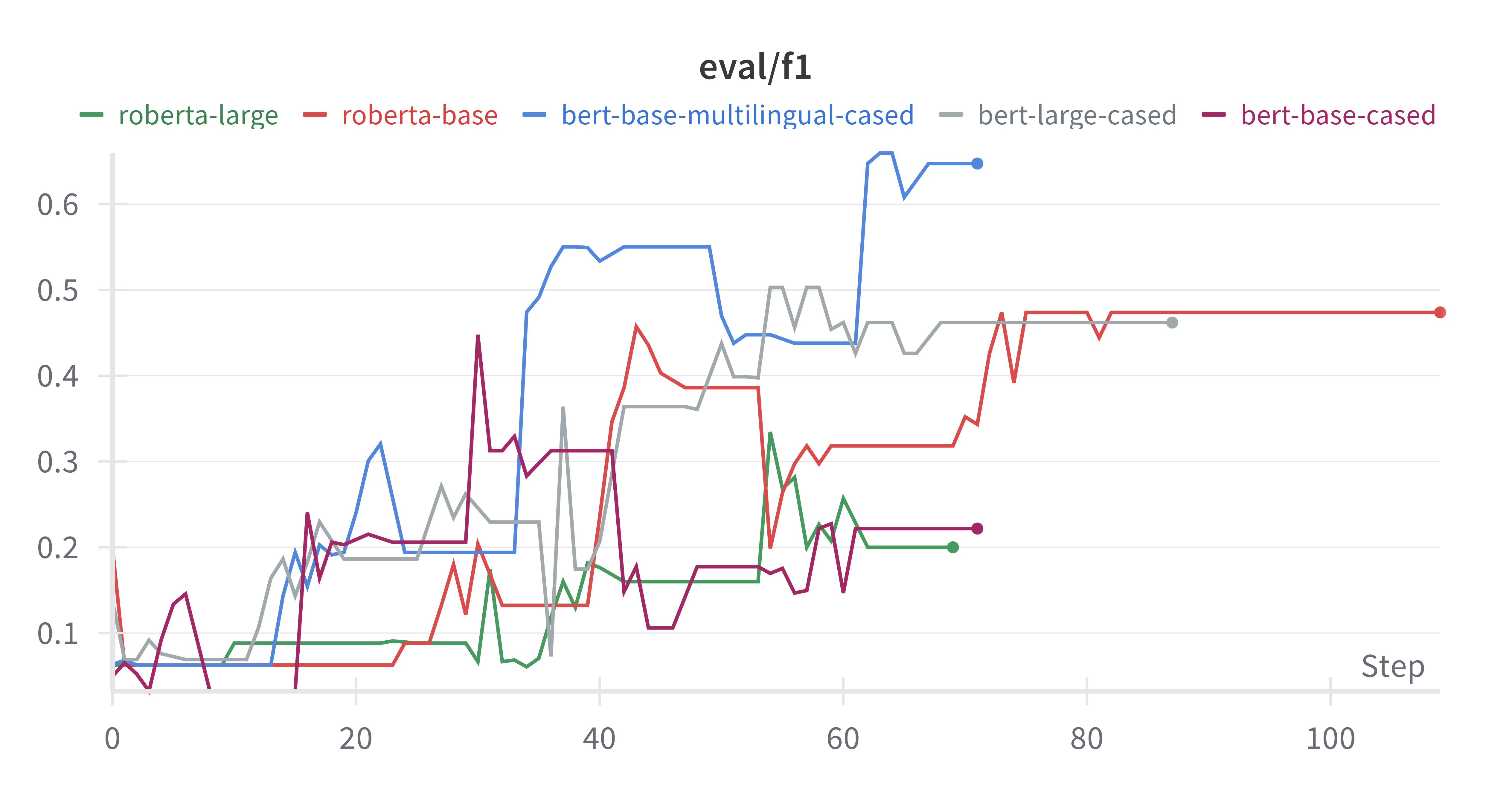}
    \caption{F1-Score Progression During Training}
    \label{fig:eval_f1}
\end{figure}

The progression of the F1-score, as shown in Figure~\ref{fig:eval_f1}, further reinforces these observations. The \texttt{mBERT} model achieved the highest F1-scores early in training and maintained steady improvement, outperforming its competitors consistently. Interestingly, increasing model size (e.g., \texttt{bert-large-cased} and \texttt{roberta-large}) did not consistently improve F1 performance, as both larger models underperformed compared to the smaller \texttt{mBERT} model. This finding suggests that model architecture and multilingual capabilities may have a more significant impact on F1 performance than size alone, underscoring the need to tailor models to the specific requirements of multilingual simplification tasks.

The \texttt{mBERT} model’s performance aligns seamlessly with the project’s primary aim of fostering multilingual accessibility, underscoring the critical importance of leveraging multilingual models to address diverse linguistic contexts and ensure inclusivity in simplification strategies.

\begin{table*}[t]
\caption{Classification Report for Typology Prediction \label{tabClassification}}
\centering
\small
\begin{tabular}{l|c|c|c|c|c|c|c}
\textbf{Class} & \multicolumn{3}{c|}{\textbf{bert-large-cased}} & \multicolumn{3}{c|}{\textbf{bert-base-multilingual-cased}} & \textbf{Support} \\
               & Precision & Recall & F1-Score & Precision & Recall & F1-Score & \\
\hline
Explanation          & 0.67 & 0.50 & 0.57 & 1.00 & 0.88 & 0.93 & 8 \\
Grammatical Adjustments & 0.00 & 0.00 & 0.00 & 0.00 & 0.00 & 0.00 & 4 \\
Modulation           & 1.00 & 0.33 & 0.50 & 0.00 & 0.00 & 0.00 & 3 \\
Omission             & 0.50 & 0.50 & 0.50 & 0.80 & 1.00 & 0.89 & 4 \\
Substitution         & 0.46 & 1.00 & 0.63 & 0.50 & 1.00 & 0.67 & 6 \\
Syntactic Changes    & 0.50 & 1.00 & 0.67 & 1.00 & 0.67 & 0.80 & 3 \\
Transposition        & 0.00 & 0.00 & 0.00 & 1.00 & 1.00 & 1.00 & 2 \\
\hline

\textbf{Avg (Macro)} & 0.45 & 0.48 & 0.47 & 0.62 & 0.70 & 0.65 & \\
\textbf{Avg (Weighted)} & 0.48 & 0.53 & 0.50 & 0.68 & 0.75 & \textbf{0.72} & \\
\hline
\textbf{Accuracy} & \multicolumn{3}{c|}{\textit{0.53}} & \multicolumn{3}{c|}{\textbf{0.70}} & 34 \\
\hline
\textbf{Training Time (s)} & \multicolumn{3}{c|}{395.22} & \multicolumn{3}{c|}{\textit{300.55}} & \\
\hline
\multicolumn{1}{c|}{\textbf{Class}} & \multicolumn{3}{c|}{\textbf{roberta-base}} & \multicolumn{3}{c|}{\textbf{roberta-large}} & \textbf{Support} \\
               & Precision & Recall & F1-Score & Precision & Recall & F1-Score & \\
\hline
Explanation          & 1.00 & 0.50 & 0.67 & 0.00 & 0.00 & 0.00 & 8 \\
Grammatical Adjustments & 0.00 & 0.00 & 0.00 & 0.00 & 0.00 & 0.00 & 4 \\
Modulation           & 1.00 & 0.33 & 0.50 & 1.00 & 0.67 & 0.80 & 3 \\
Omission             & 0.75 & 0.75 & 0.75 & 1.00 & 0.25 & 0.40 & 4 \\
Substitution         & 0.43 & 1.00 & 0.60 & 0.25 & 0.40 & 0.31 & 6 \\
Syntactic Changes    & 0.60 & 1.00 & 0.75 & 0.67 & 0.67 & 0.67 & 3 \\
Transposition        & 0.25 & 0.50 & 0.33 & 0.00 & 0.00 & 0.00 & 2 \\
\hline

\textbf{Avg (Macro)} & 0.47 & 0.51 & 0.48 & 0.28 & 0.28 & 0.27 & \\
\textbf{Avg (Weighted)} & 0.50 & 0.53 & \textit{0.52} & 0.30 & 0.35 & 0.32 & \\
\hline
\textbf{Accuracy} & \multicolumn{3}{c|}{\textit{0.53}} & \multicolumn{3}{c|}{0.30} & 34 \\
\hline
\textbf{Training Time (s)} & \multicolumn{3}{c|}{\textbf{219.30}} & \multicolumn{3}{c|}{587.21} & \\
\hline
\end{tabular}
\end{table*}

\section{Interpretability of predictions}
\label{sec:org0d8d753}
We have trained a classifier for predicting the difficulty of sentences by means of collecting simple and difficult sentences from Wikipedia and fine-tuning mBERT \cite{devlin2019bert}.

By means of the implementation of the Integrated Gradients in the Captum library \cite{miglani23captum}, we can:
\begin{enumerate}
\item detect which words or syntactic constructions commonly affect readability, as well as
\item which of them align with human annotation.
\end{enumerate}
We utilised the Integrated Gradients (IG) method to identify the tokens in a sentence that contributed most significantly to the model's predictions. IG achieves this by calculating the gradients of the model's output with respect to its input, thereby highlighting the importance of individual features.

\noindent \textbf{For Example:} Consider the following sentence from our dataset:
\begin{quote}
\textit{``Provide financially sustainable care, giving security and stability to people and their carers.''}
\end{quote}

The Integrated Gradients approach offered actionable insights by attributing importance scores to specific words, revealing their influence on the model's predictions. For this sentence, the prediction probabilities are: \textbf{Simple:} 0.02, and  \textbf{Complex:} 0.98.

\begin{itemize}
    \item \textbf{High-impact words:} The IG method highlighted domain-specific and content-heavy words such as \textit{``sustainable,'' ``security,'' and ``stability''}, which were crucial for determining that the sentence was \textit{``Complex.''}
    \item \textbf{Stopwords:} Words with minimal semantic content (e.g., \textit{``and,'' ``to,'' ``their''}) were assigned near-zero attribution scores, as expected.
    \item \textbf{Prediction Analysis:} Based on the probabilities, the sentence was classified as \textit{Complex} with a high confidence of 98\%.
\end{itemize}
\begin{table}[t]
\caption{Word-level Attributions for the Example Sentence \label{tabAttributions}}

\small
\begin{tabular}{l|r|r}
\textbf{Word} & \textbf{Attribution } & \textbf{Contribution} \\[0pt]
\hline
Provide & 0.18 & Moderately Complex \\[0pt]
financially & -0.10 & Slightly Easy \\[0pt]
sustainable & 0.30 & Highly Complex \\[0pt]
care & 0.15 & Slightly Complex \\[0pt]
giving & 0.10 & Slightly Complex \\[0pt]
security & 0.25 & Highly Complex \\[0pt]
and & -0.02 & Neutral \\[0pt]
stability & 0.28 & Highly Complex \\[0pt]
to & -0.03 & Neutral \\[0pt]
people & 0.12 & Slightly Complex \\[0pt]
and & -0.04 & Neutral \\[0pt]
their & 0.05 & Neutral \\[0pt]
carers & -0.08 & Neutral \\[0pt]
\end{tabular}
\end{table}

\begin{figure}[h!]
    \centering
    \includegraphics[width=0.5\textwidth]{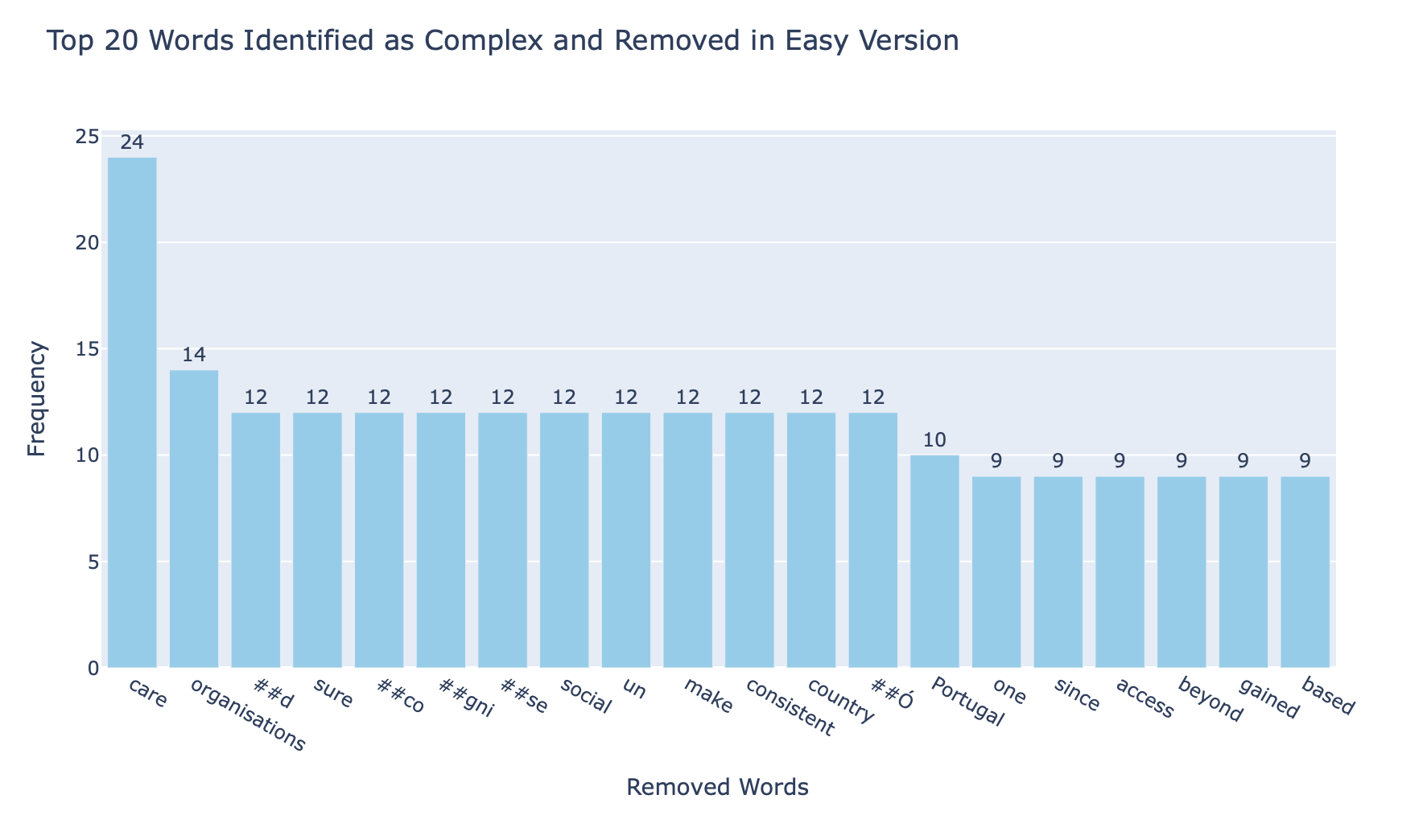} 
    \caption{Top 20 Words Identified as Complex and Removed in Easy Version}
    \label{fig:removed_words}
\end{figure}

 By applying the IG method, we identified a total of 1303 complex words from the original sentences. These words were then compared against their corresponding simplified, E2R versions to determine which complex words were removed during simplification. This comparison yielded 877 removed words, representing 67.31\% of the total complex words identified. The removed words are indicative of tokens that were deemed complex by both the model and human editors, as their removal from the E2R versions suggests that they were perceived as difficult or unnecessary for simplified comprehension. This alignment between the model-predicted complex words and those removed in human-curated simplifications demonstrates the model's effectiveness in predicting words that are likely to be complex and corroborates the utility of the IG method for interpretability in text simplification tasks. As shown in \textbf{Figure~\ref{fig:removed_words}}, the most frequently removed complex words included meaningful content terms such as \textit{"care," "organisations,"} and \textit{"consistent."}
 % highlighting the close connection between linguistic complexity and simplification practices.

\section{Findings and Contributions}

The findings demonstrate that transformer-based models are capable of handling the complexities of typology classification, especially when supported by preprocessing techniques and loss weighting strategies. The model exhibits moderate success in identifying phenomena that require simplification. However, it encounters notable challenges with underrepresented classes and specific simplification strategies, such as \textit{``grammatical adjustments''} and \textit{``omission.''}
% which are significantly impacted by class imbalance. 

In summary, while transformer-based models hold considerable potential for simplifying texts to improve accessibility, addressing class imbalance through the use of comprehensive, balanced datasets is crucial. Leveraging the complete dataset further enhances the model’s reliability and enables it to generalise effectively across all simplification categories.

One of the critical findings of this study is the utility of the IG framework for interpretability. IG provides insights that align closely with human annotations regarding complexity. For example, IG effectively identifies tokens contributing to difficulty, such as \textit{``sustainable''} or \textit{``stability''}, while assigning minimal importance to semantically neutral words like \textit{``and''} or \textit{``to.''} This alignment bridges the gap between machine predictions and human reasoning, enabling iterative improvements in model development.

The alignment of the model's predictions with the removal of complex words by human editors demonstrates its capability to predict readability effectively. In particular, 67.31\% of the complex words identified by IG were removed in the human-simplified versions, highlighting the model's predictive accuracy in real-world applications.

Moreover, the study shows the close connection between linguistic complexity and simplification practices. Frequent removal of meaningful content words, such as \textit{``care,'' ``organisations,''} and \textit{``consistent,''} highlights the importance of meaning and context in making texts easier to understand for different audiences.

\section{Conclusions}
\label{sec:orge95f00f}

Building on the annotation framework, several key insights emerge regarding the challenges and strategies involved in translating texts into E2R English. First, intralingual translation facilitates a more straightforward comparison between source and target texts due to the inherent isomorphism between the source and target languages. Second, the choice of translation strategies must be tailored to the specific type of intralingual translation, ensuring that the target text aligns with its intended function. For example, in diastratic translation—specifically the transformation of standard English into E2R English—the focus lies on simplifying vocabulary, syntax, and semantic structures while maintaining fidelity to the source text and accessibility for the target audience.

Moreover, the proposed taxonomy, encompassing 9 macro-strategies, 33 strategies, and 15 micro-strategies, illustrates the cognitive complexity of intralingual translation. These challenges underscore the limitations of current automation tools, as computational analyses reveal the nuanced skills required for transcription and modification strategies. Even in the era of generative artificial intelligence, text simplification remains a non-trivial task due to its intricate linguistic demands.

The novelty of our approach lies not only in the dataset itself but also in the methodology, which bridges translation studies and text simplification by categorizing transformations into well-defined categories. This integration offers new insights into the strategies employed in simplification and provides a robust framework for developing models that can generalise across multiple types of linguistic transformations.

The results highlight the significant progress achieved with our approach, as the fine-tuned \texttt{mBERT} model outperformed the baseline majority-class strategy, which achieved an accuracy of 24.5\% and a weighted F1-score of 9.6\%. In contrast, \texttt{mBERT} achieved 70\% accuracy, a weighted F1-score of 72\%, and a macro F1-score of 65\%, demonstrating its ability to generalise across majority and minority classes. 
% It excelled particularly in critical categories like \textit{Explanation} (F1: 0.93), \textit{Substitution} (F1: 0.67), and \textit{Syntactic Changes} (F1: 0.80).

%% The results demonstrate that \texttt{mBERT} offers an optimal balance between performance and computational efficiency. With a short training time significantly outperforming larger models like \texttt{roberta-large} in both accuracy and F1-scores. This highlights that increasing model size does not necessarily translate to better performance, and smaller, well-optimised models like \texttt{mBERT} can deliver superior results more efficiently, making them a more practical choice for resource-constrained settings.

Employing Integrated Gradients (IG) enhances the interpretability of model predictions, ensuring closer alignment with human annotations. IG offers a clearer understanding of the input data elements the model prioritises, thereby elucidating its decision-making processes. Our primary results align with the identification of complex words that were either modified or removed in the simplified versions. In particular, 67.31\% of the complex words identified by IG were removed in the human-simplified versions, highlighting the model’s accuracy in applications. This transparency is critical for identifying strengths and weaknesses, guiding iterative improvements, and fostering trust in machine-generated outputs. Additionally, IG serves as a tool to validate the predictions of the LLM model against expert judgments, ensuring reliability and consistency in its reasoning, and ensuring that it makes the right predictions for the right reasons \cite{schramowski20right}.

Future research should prioritise addressing class imbalance through advanced techniques such as hierarchical annotations, domain-specific embeddings, or data augmentation. Incorporating multiple annotators would also enable the calculation of agreement metrics, improving the evaluation of annotation reliability. Expanding the interpretability framework to cross-linguistic simplifications presents another promising avenue. Leveraging the full Scottish Government dataset and employing advanced machine learning techniques could further enhance performance across all linguistic categories. This work ultimately contributes to the broader goal of creating accessible, inclusive texts while promoting trust and transparency in AI-driven systems.

\section*{Acknowledgments}

This document is part of a project that has received funding from the European Union’s Horizon Europe research and innovation program under the Grant Agreement No. 101132431 (iDEM Project). The views and opinions expressed in this document are solely those of the author(s) and do not necessarily reflect the views of the European Union. Neither the European Union nor the granting authority can be held responsible for them.

The University of Leeds (UOL) was funded by \textbf{UK Research and Innovation (UKRI)} under the UK government’s Horizon Europe funding guarantee (Grant Agreement No. 10103529).

\bibliography{bibexport}

\end{document}